# Technical Report:
# Applying the Lower-Biased Teacher Model
# in Semi-Supervised Object Detection

Author: Shuang Wang

Gustavo Batista

# Abstract

In this research report, I present the Lower Biased Teacher model, an enhancement of the Unbiased Teacher model, specifically tailored for semi-supervised object detection tasks. The primary innovation of this model is the integration of a localization loss into the teacher model, which significantly improves the accuracy of pseudo-label generation. By addressing key issues such as class imbalance and the precision of bounding boxes, the Lower Biased Teacher model demonstrates superior performance in object detection tasks. Extensive experiments on multiple semi-supervised object detection datasets show that the Lower Biased Teacher model not only reduces the pseudo-labeling bias caused by class imbalances but also mitigates errors arising from incorrect bounding boxes. As a result, the model achieves higher mAP scores and more reliable detection outcomes compared to existing methods. This research underscores the importance of accurate pseudo-label generation and provides a robust framework for future advancements in semi-supervised learning for object detection.

## 1. Introduction

### 1.1 Background

With the advancement of deep learning, several models have demonstrated remarkable performance in image classification and object detection tasks. However, to fully exploit the capabilities of deep learning models, extensive labeled datasets are essential, which involve significant time and cost for labeling. As an alternative, semi-supervised learning has gained increasing attention in recent years (Sohn et al., 2020). Nevertheless, most research on semi-supervised learning has concentrated on image classification tasks rather than object detection, which requires a larger number of labeled bounding boxes. This research proposes a semi-supervised learning approach using both labeled and mostly unlabeled data. Current popular semi-supervised learning methods face some challenges: dataset imbalances can affect the quality of pseudo labels (Lee, 2013); some images lack clear differentiation between foreground and background, resulting in less accurate localization of predicted bounding box positions; more importantly, existing semi-supervised learning methods that perform well in classification often do not produce satisfactory results in object detection (Sohn et al., 2020). This discrepancy arises because classification tasks aim to categorize input data into predefined classes by identifying key patterns and features, which is relatively straightforward. Semi-supervised learning models can effectively utilize unlabeled data to enhance this mapping process (i.e., from inputs to categories). However, object detection tasks are more complex as they require determining the precise locations of objects, typically represented by bounding boxes, before further identification. Hence, object detection is a more intricate task. Thus, object detection is a complex task with additional challenges arising from several factors:

**Variation in image size**: Objects can appear in multiple sizes, poses, and angles.

**Occlusion and interaction**: Objects may occlude each other or interact in complex ways, making detection more difficult.

**Complexity of background**: Backgrounds in object detection tasks can be highly intricate, containing distracting objects or patterns that can mislead semi-supervised learning models.

**Demand for precise bounding box annotation**: Detection tasks require more precise annotations, such as bounding boxes, which are often not available in semi-supervised learning settings (Ren et al., 2015).

Given the mentioned reasons, this research proposes a lower-biased teacher model that can generate more accurate pseudo labels, implement a better weight updating method, and build a more precise bounding box to enhance the model's generalization ability and reduce overfitting in object detection tasks.

**1.2 Dataset**

**COCO (Common Objects in Context) Dataset**

The COCO dataset is a large-scale object detection, segmentation, and captioning dataset. COCO has several features that make it an advanced benchmark for these types of tasks. It includes images of complex everyday scenes containing common objects in their natural context. Objects are labeled using precise segmentation masks to enable pixel-accurate segmentation. The dataset contains photos of 91 object types that would be easily recognizable by a 4-year-old child. With over 2.5 million labeled instances in over 328,000 images, the dataset is widely used to train deep learning models that can understand images and provide detailed object-level annotations. Additionally, COCO provides multiple challenges such as object detection, segmentation, key point detection, panoptic segmentation, and image captioning, making it versatile for various vision-based tasks.

**PASCAL VOC (Visual Object Classes) Dataset**

The PASCAL VOC dataset is a historic dataset in computer vision used for object detection and image classification. It was launched in the 2005 PASCAL Visual Object Classes Challenge. Over the years, it has been a benchmark for image classification, object detection, and segmentation tasks. The dataset includes images from a wide range of categories such as animals, vehicles, household objects, and people, annotated for various tasks including classification, detection, and segmentation. The dataset is smaller than COCO, consisting of tens of thousands of images and covering 20 different object classes. PASCAL VOC has been fundamental in advancing object detection algorithms and is known for its well-defined and clear evaluation protocols, although it is less frequently used today due to the rise of larger, more complex datasets like COCO.

## 2. Literature Review

Traditional semi-supervised learning methods mainly include generative models, self-training models, entropy-consistency regularization models, and graph-theory-based models (Verma et al., 2022). The entropy-consistency regularization approach, which can be easily implemented through loss functions, has gradually become a popular direction in deep semi-supervised learning research. Early research by (Miller et al. 1996) theoretically demonstrated the potential of using unlabeled data to improve model classification performance, providing intuitive analysis from the perspective of data distribution estimation. Current studies have shown that training with a small amount of labeled data using semi-supervised methods can achieve performance comparable to fully supervised learning with complete labeled data within the same dataset. This provides theoretical support for the research methods discussed in this proposal.

**2.1 Pseudo label**

Initially, all the labeled data are used to train an initial model, which will then predict the labels of the unlabeled data. These predicted labels are referred to as 'pseudo labels,' although they may not be very accurate (Wang et al., 2018). Typically, pseudo labels are selected only when the model has high confidence in its predictions, which is managed through a confidence threshold setting (Xie et al., 2021). The quality of pseudo labels is crucial to the performance of the models. Inaccurate pseudo labels can lead to poor performance. During iterative processing, dynamically adjusting the confidence threshold or altering the strategy for selecting pseudo labels may improve model performance.

**2.2 Consistency Regularization**

In deep learning, consistency regularization is widely used to manage model complexity and prevent overfitting (Jeong et al., 2019). Unlike traditional regularization techniques like L1 and L2, which use penal parameters to reduce complexity, consistency regularization focuses on maintaining output smoothness across different inputs. It involves an unsupervised component that predicts the output for different inputs and ensures output consistency. The similarity degree is maintained by the penalty parameters in the loss function, similar to L1 and L2 distances but using more complex algorithms such as kernel correlation. This method is widely applied in deep learning, effectively enhancing the generalization ability of models without significantly increasing computational demands.

**2.3 Mean Teacher**

In semi-supervised learning, the Mean Teacher framework includes the 'Teacher model' and the 'Student model.' The student model serves as the training model, while the teacher model is a smoothed version of the student model. The student model updates its weights using standard backpropagation during training, whereas the teacher model's weights are updated as a moving average of the student model's weights (Wang et al., 2023). Thus, the teacher model's weights represent the average historical weights of the student model. This process is illustrated in equation (1).

$$\theta_t = \alpha\theta_t + (1-\alpha)\theta_s \qquad (1)$$

In this equation, α is used to control the update rate, $\theta_t$ is the parameter of the teacher model, and $\theta_s$ is the parameter of the student model. The Mean Teacher method enhances learning by making predictions on unlabeled data. The system generates predictions for these unlabeled data and then compares them with the predictions of the teacher model (Han et al., 2019). This comparison helps guide the learning of the student model, allowing it to generalize better to new data. The loss function of the Mean Teacher model usually consists of two parts: one part based on the supervised loss from labeled data, and the other part based on the unsupervised loss generated from the consistency between the outputs of the student and teacher models (Wang et al., 2023). This approach enables the model to learn the correct outputs for labeled data and maintain consistent predictions on unlabeled data. Therefore, the total loss is the weighted sum of the supervised and unsupervised losses.

## 3. Methodology

**3.1 Unbiased Teacher**

Unbiased Teacher (UBT) model includes two part-burn in part and mutual learning part (Liu et al. 2021). For the burn in part, they first use the use the available supervised data to optimize their model θ with the supervised loss, the supervised loss of object detection consists of four losses: the RPN classification loss, the RPN regression loss, the ROI classification loss, and the ROI regression loss. After burn-in, they introduce the Teacher-Student Mutual Learning regimen, where the student is optimized by using the pseudo-labels generated by the Teacher, and the Teacher is updated by gradually transferring the weights of the continually learned Student model. With the interaction between the Teacher and the Student, both models can evolve jointly and continuously to improve detection accuracy. An overview of UBT's model is shown in Figure 1.

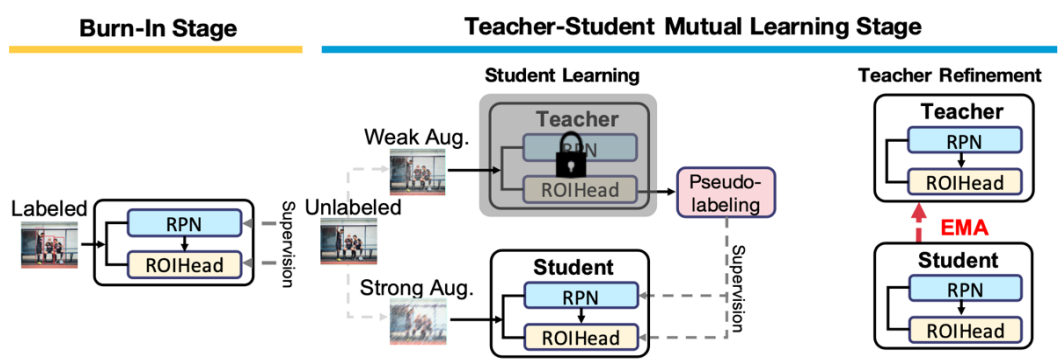

Figure 1 (Liu et al. 2021)

**3.2 Consistency-based Semi-Supervised Learning for Object Detection**

The structure of the Consistency-based Semi-Supervised Learning for Object Detection (CSD) model (Jeong et al. 2019) combines the elements of a semi-supervised learning (SSL) model and an object detection algorithm. To ensure a one-to-one correspondence of target objects, both an original image I and its flipped version are used as inputs. A paired bounding box should represent the same class, and their localization information must remain consistent. During the training process, each mini batch includes both labeled and unlabeled images. The labeled samples are trained using a typical object detection approach, while consistency loss is applied to both labeled and unlabeled images. In object detection, an additional class, 'background', exists, and most candidate boxes are usually classified into this class unless filtered by a confidence threshold. Consequently, consistency losses computed with all candidates can be easily dominated by background instances, which can degrade classification performance for the foreground classes. To address this, boxes with a high probability of being background are excluded by marking them with a mask. An overview of CSD model is shown in Figure 2.

### 3.3 Lower Biased Teacher Model

#### 3.3.1 Burn in Loss

Lower-Bias Teacher Model is a model that combines UBT and CSD models. In the initial phase of model training, commonly referred to as the "burn-in" phase, I incorporate the previously mentioned CSD method into the supervised learning section. This strategy aims to enable the model to learn more precise and robust feature representations from labeled data using both the original and flipped images. I feed both the original and flipped images into the Faster R-CNN model and incorporate the localization loss (as shown in equation 3) into the UBT model's Burn-In loss (shown as equation 4). So the total loss $L_{burn\_in}$ is Faster-RCNN's original loss plus localization loss (shown as equation 5).

The localization result for the k-th candidate box $f_{loc}^k(I)$ consists of [Δcx, Δcy, Δω, Δh], which represent the displacement of the center and scale coefficients of a candidate box, respectively. $f_{loc}^k(I)$ and $f_{loc}^{k'}(I')$ require a simple modification to be equivalent to each other. Since the flipping transformation makes $\Delta c'x'$ move in the opposite direction, a negation should be applied to correct it. This approach aims to train more accurate teacher and student models.

$$l_{con\_loc}\left(f_{loc}^k(I), f_{loc}^{k'}(I')\right) = \frac{1}{4}(||\Delta cx^k - (-\Delta c'x'^{k'})||^2 + ||\Delta cy^k - \Delta c'y'^{k'}||^2$$

$$+||\Delta \omega^k - \Delta \omega'^{k'}||^2 + ||\Delta h^k - \Delta h'^{k'}||^2) \quad (3)$$

$$L_{UBT\_burn\_in} = L_{cls}^{rpn} + L_{reg}^{rpn} + L_{cls}^{roi} + L_{reg}^{roi} \quad \text{(Liu et al.)} \quad (4)$$

$$L_{burn\_in} = L_{UBT\_burn\_in} + L_{con\_loc} \quad (5)$$

#### 3.3.2 Mutual Learning

To leverage unsupervised data, I introduce the Teacher-Student Mutual Learning

regimen. In this approach, the Student is optimized using the pseudo-labels generated by the Teacher, and the Teacher is updated by gradually transferring the weights from the continually learned Student model. With the interaction between the Teacher and the Student, both models evolve jointly and continuously, improving detection accuracy. This improvement means that the Teacher generates more accurate and stable pseudo-labels, which is key to achieving significant performance gains compared to existing methods. The Teacher can also be regarded as the temporal ensemble of the Student models at different time steps, explaining why the Teacher's accuracy consistently surpasses that of the Student. To enhance the Teacher model (Tarvainen et al. 2017), I use strongly augmented images as input for the Student and weakly augmented images for the Teacher, ensuring reliable pseudo-labels.

To address the lack of ground-truth labels for unsupervised data, I adapt the pseudo-labeling method to generate labels for training the Student with unsupervised data. This follows the principle of successful examples in semi-supervised image classification tasks. Similar to classification-based methods, to prevent the detrimental effects of noisy pseudo-labels (i.e., confirmation bias or error accumulation), I first set a confidence threshold $\delta$ for predicted bounding boxes to filter out low-confidence predictions, which are more likely to be false positives.

While the confidence threshold method has achieved tremendous success in image classification, it is not sufficient for object detection. This is because duplicated box predictions and imbalanced prediction issues also exist in semi-supervised object detection. To address the duplicated box prediction issue, I remove repetitive predictions by applying class-wise non-maximum suppression (NMS) before using confidence thresholding, as performed in STAC (Sohn et al. 2020b).

To obtain more stable pseudo-labels, I apply EMA to gradually update the Teacher model. The slowly progressing Teacher model can be regarded as the ensemble of the Student models in different training iterations. For eliminate bias in pseudo-labels, I use the focal loss rather than cross-entropy loss for multi-class classification of ROIhead classifier to focus on hard samples or not domination samples. Besides, I add the Consistency Localization Loss to the unbiased teacher's supervised loss. The purpose of this loss function is to enhance the model's generalization ability on unlabeled data by enforcing spatial consistency between the teacher and student models. This improves performance on unlabeled data and makes the teacher model more unbiased.

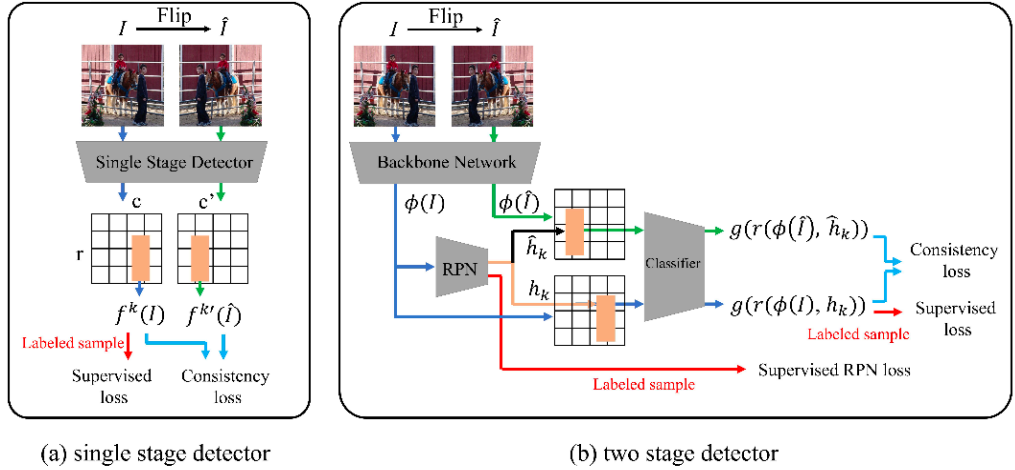

Figure 2 (Jeong et al. 2019)

## 4. Experiments

I benchmark my proposed method on experimental settings using MS-COCO and PASCAL VOC following existing works (Everingham et al. 2010). Specifically, there are two experimental settings: (1) COCO-standard: I randomly sample 0.5, 1, 2, 5, and 10% of labeled training data as a labeled set and use the rest of the data as the training unlabeled set. (2) VOC: I use the VOC07 trainval set (5001 pieces of image) as the labeled training set and the VOC12 trainval set (11540 pieces of image) as the unlabeled training set. Model performance is evaluated on the VOC07 test set.

For a fair comparison, I followed the Unbiased Teacher paper to use Faster- RCNN with FPN as my object detector, where the feature weights are initialized by the ImageNet-pretrained model. I use the confidence threshold δ = 0.7. For the data augmentation, I apply random horizontal flips for weak augmentation and randomly add color jittering, grayscale, Gaussian blur, and cutout patches for strong augmentations. And I do not apply any geometric augmentations. I use AP50 and AP50:95 (denoted as mAP) as evaluation metric by the prior work (Cai et al. 2018), and the performance is evaluated on the Teacher model.

**Training Iterations:** For Lower biased Teacher model and CSD and UBT model, I all use the same (27k) iterations for training.

**Hardware Information**: Four RTX A5000 24G GPUs are used to train the CSD model. Two RTX A800 80G GPUs and five RTX A6000 48G GPUs are used to train the UBT model. Four RTX A800 80G GPUs and six RTX A6000 48G GPUs are used to train the

Lower Biased Teacher model. Since these models are trained on different hardware clusters, the use of more GPUs may result in higher accuracy.

## 5. Result

**COCO:** I first evaluate the efficacy of my Lower Biased Teacher on COCO-standard (Table 1). When there are only 0.5% to 10% of data labeled, our model consistently performs favorably against the UBT model's method (Liu et al. 2021) methods, and CSD (Jeong et al., 2019). I also observe that, as the amount of labeled data decreases, the improvements of my method over existing approaches become more significant. The Unbiased Teacher consistently shows around 1-2 absolute mAP improvements when using less than 5% of labeled data compared to the UBT model's method. I attribute these improvements to several crucial factors:

**More accurate pseudo-labels**: When adding localization loss to the previous model (Liu et al., 2021) and leveraging the pseudo-labeling and consistency regularization between the two networks (Teacher and Student in my case), it is critical to ensure that the pseudo-labels are accurate and reliable. Existing methods attempt to do this by training the pseudo-label generation model using all the available labeled data and then freezing it completely. In contrast, in my framework, the pseudo-label generation model (Teacher) continues to evolve gradually and smoothly through Teacher-Student Mutual Learning. This enables the Teacher to generate more accurate pseudo-labels, which are effectively utilized in the training of the student.

Table 1: Experimental results on COCO-standard comparing with CSD (Jeong et al., 2019) and UBT (Liu et al., 2021).

|      | COCO Standard | | | | |
| --- | --- | --- | --- | --- | --- |
|      | 0.50% | 1.00% | 2% | 5% | 10% |
| CSD  | 7.4±0.3 | 10.1±0.1 | 12.5±0.1 | 18±0.2 | 24±0.1 |
| UBT  | 17.1±0.3 | 20.51±0.2 | 24.2±0.27 | 28.32±0.15 | 31.5±0.1 |
| Ours | 19.1±0.1 | 22.12±0.1 | 25.1±0.15 | 29.81±0.21 | 32.1±0.2 |

**VOC:** In the previous section, I demonstrated that the Unbiased Teacher can successfully leverage very small amounts of labeled data. Now, I aim to verify whether the model trained on more supervised data can be further improved by using additional unlabeled data. Therefore, To further examine whether increasing the size of unlabeled data can further improve the performance, I follow CSD and UBT to use COCO20cls

dataset3 (random chose 5k pieces of image) as an additional unlabeled set. And the result shows in Table 2.

Table2: Results on VOC comparing with CSD (Jeongetal.,2019) and UBT (Liu et al. 2020b).

|  | Labeled | Unlabeled | AP50 | AP50:95 |
| --- | --- | --- | --- | --- |
| CSD | VOC07 | VOC12 | 74.4±0.3 | - |
| UBT |  |  | 76.71±0.18 | 48.9±0.2 |
| Ours |  |  | 77.95±0.2 | 51.11±0.1 |
|  |  |  |  |  |
| CSD |  | VOC12 | 75.0±0.1 | - |
| UBT | VOC07 | + | 77.97±0.21 | 50.42±0.1 |
| Ours |  | COCO20cls | 79.15±0.82 | 52.08±0.12 |

## 6. Conclusion

In this research, I revisit the semi-supervised object detection task. By analyzing object detectors in scenarios with limited labeled data, I identify and address two major issues: accuracy and class imbalance. I propose the Lower Biased Teacher—a framework where a Teacher with localization constraints and a Student learn jointly to improve each other. In the experiments, I demonstrate that my model mitigates the pseudo-labeling bias caused by class imbalance and reduces errors from incorrect bounding boxes. The Lower Biased Teacher achieves satisfactory performance across multiple semi-supervised object detection datasets.

Table2: Results on VOC comparing with CSD (Jeongetal.,2019) and UBT (Liu et al. 2020b).